\documentclass[runningheads]{llncs}

\usepackage{epsfig}
\usepackage{subfigure}
\usepackage{wrapfig}
\usepackage{url}
\usepackage{enumerate}

\begin{document}

\pagestyle{headings}
\mainmatter

\title{Parameter-less Hierarchical BOA}
\titlerunning{Parameter-less Hierarchical BOA}

\author{Martin Pelikan and Tz-Kai Lin}
\authorrunning{Martin Pelikan and Tz-Kai Lin}
\authorrunning{Martin Pelikan and Tz-Kai Lin}

\institute{Dept. of Math. and Computer Science, 320 CCB\\
University of Missouri at St. Louis\\
8001 Natural Bridge Rd.,
St. Louis, MO 63121\\
\email{pelikan@cs.umsl.edu}\\
\email{tlkq4@studentmail.umsl.edu}
}


\maketitle 


\begin{abstract}
The parameter-less hierarchical Bayesian optimization algorithm (hBOA) enables the use of hBOA without the need for tuning parameters for solving each problem instance. There are three crucial parameters in hBOA: (1) the selection pressure, (2) the window size for restricted tournaments, and (3) the population size. Although both the selection pressure and the window size influence hBOA performance, performance should remain low-order polynomial with standard choices of these two parameters. However, there is no standard population size that would work for all problems of interest and the population size must thus be eliminated in a different way. To eliminate the population size, the parameter-less hBOA adopts the population-sizing technique of the parameter-less genetic algorithm. Based on the existing theory, the parameter-less hBOA should be able to solve nearly decomposable and hierarchical problems in quadratic or subquadratic number of function evaluations without the need for setting any parameters whatsoever. A number of experiments are presented to verify scalability of the parameter-less hBOA. 
\end{abstract}


\section{Introduction}
When the hierarchical Bayesian optimization algorithm (hBOA) was designed~\cite{Pelikan:01*,Pelikan:03b}, it was argued that hBOA can solve difficult nearly decomposable and hierarchical problems without the need for setting any parameters. As a result, to solve a new black-box optimization problem, it should be sufficient to plug the new problem into hBOA, press the start button, and wait until hBOA figures out where the optimum is. It was argued that all hBOA parameters except for the population size can be set to their default values without affecting good scalability of hBOA~\cite{Pelikan:02a}. However, choosing an adequate population size was argued to be crucial~\cite{Pelikan:02a} and in all experiments the user was assumed to set the population size optimally to obtain best performance while retaining reliable convergence. That is why the dream of having a fully parameter-less optimizer for the entire class of nearly decomposable and hierarchical problems remained one parameter away from reality.

The purpose of this paper is to propose a fully parameter-less hBOA by implementing the parameter-less population-sizing scheme of the parameter-less genetic algorithm (GA)~\cite{Harik:99a} in hBOA. The parameter-less hBOA simulates a collection of populations of different sizes, starting with a small base population size. Each next population is twice as large as the previous one. To enable parallel simulation of a number of populations when the number of populations that must be simulated is not known in advance, each population is forced to proceed with at most the same speed as any smaller population, where speed is considered with respect to the number of function evaluations. This ensures that although there is no upper bound on the number of populations simulated in parallel and on their size, the overall computational overhead is still reasonable compared to the case with an optimal population size. In fact, theory exists that shows that the parameter-less population sizing scheme in the parameter-less GA does not increase the number of function evaluations until convergence by more than a logarithmic factor~\cite{Pelikan:99c}. As a result, hBOA can be expected to perform within a logarithmic factor from the case with the optimal population size. This is verified with a number of experiments on nearly decomposable and hierarchical problems. 

The paper starts by discussing probabilistic model-building genetic algorithms (PMBGAs) and the hierarchical BOA (hBOA). Section~\ref{section-plGA} discusses the parameter-less genetic algorithm, which serves as the primary source of inspiration for designing the parameter-less hBOA. Section~\ref{section-parameter-less-hBOA} describes the parameter-less hBOA. Section~\ref{section-experiments} describes experiments performed and discusses empirical results. Finally, Section~\ref{section-conclusions} summarizes and concludes the paper.  


\section{Hierarchical Bayesian optimization algorithm (hBOA)}
\label{section-hBOA}

Probabilistic model-building genetic algorithms (PMBGAs)~\cite{Pelikan:02,Larranaga:02} replace traditional variation operators of genetic and evolutionary algorithms by a two-step procedure. In the first step, a probabilistic model is built for promising solutions after selection. Next, the probabilistic model is sampled to generate new solutions. By replacing variation operators inspired by genetics with machine learning techniques that allow automatic discovery of problem regularities from populations of promising solutions, PMBGAs provide quick, accurate, and reliable solution to broad classes of difficult problems, many of which are intractable using other optimizers~\cite{Pelikan:01*,Pelikan:03b}.

For an overview of PMBGAs, please see references~\cite{Pelikan:02} and~\cite{Larranaga:02}. PMBGAs are also known as estimation of distribution algorithms (EDAs)~\cite{Muhlenbein:96**} and iterated density-estimation algorithms (IDEAs)~\cite{Bosman:00*}. The remainder of this section describes the hierarchical Bayesian optimization algorithm, which is one of the most advanced and powerful PMBGAs. 

\subsection{Hierarchical BOA (hBOA)}
The hierarchical Bayesian optimization algorithm (hBOA)~\cite{Pelikan:01*,Pelikan:03b} evolves a population of candidate solutions to the given problem. The first population of candidate solutions is usually generated at random. The population is updated for a number of iterations using two basic operators: (1) selection, and (2) variation. The selection operator selects better solutions at the expense of the worse ones from the current population, yielding a population of promising candidates. The variation operator starts by learning a probabilistic model of the selected solutions. hBOA uses Bayesian networks with local structures~\cite{Chickering:97} to model promising solutions. The variation operator then proceeds by sampling the probabilistic model to generate new solutions, which are incorporated into the original population using the restricted tournament replacement (RTR)~\cite{Harik:95a}. RTR ensures that useful diversity in the population is maintained for long periods of time. The run is terminated when a good enough solution has been found, when the population has not improved for a long time, or when the number of generations has exceeded a given upper bound. Figure~\ref{figure-hboa} shows the pseudocode of hBOA. For a more detailed description of hBOA, see~\cite{Pelikan:thesis}.

\begin{figure}[t]
\begin{verbatim}
Hierarchical BOA (hBOA)
  t := 0;
  generate initial population P(0);
  while (not done) {
    select population of promising solutions S(t);
    build Bayesian network B(t) with local struct. for S(t);
    sample B(t) to generate offspring O(t);
    incorporate O(t) into P(t) using RTR yielding P(t+1);
    t := t+1;
  };
\end{verbatim}
\caption{The pseudocode of the hierarchical BOA (hBOA).}
\label{figure-hboa}
\end{figure}

\section{Parameter-less genetic algorithm}
\label{section-plGA}

The parameter-less genetic algorithm (GA)~\cite{Harik:99a} eliminates the need for setting parameters---such as the population size, selection pressure, crossover rate, and mutation rate---in genetic algorithms. The crossover rate and selection pressure are set to ensure consistent growth of building blocks based on the schema theorem. The mutation rate can be eliminated in a similar manner. The population size is eliminated by simulating a collection of populations of different sizes. In the context of hBOA, eliminating the population size is the most important part of the parameter-less GA, because the selection pressure influences performance of hBOA by only a constant factor and there are no crossover or mutation rates in hBOA. 

\subsection{Eliminating selection pressure and crossover rate}
The parameter-less GA assumes that selection and recombination are primary search operators. The choice of selection pressure and crossover rate should consider the following facts. Selection must be strong enough to ensure consistent growth of superior building blocks of the optimum, but it must not be too strong because otherwise diversity in the population might be lost prematurely. On the other hand, the crossover rate must be large enough to ensure sufficient exploration of the search space but if blind crossover operators (e.g., one-point and uniform crossover) are used, each application of crossover can break an important building block and thus crossover must not be applied too frequently. 

Additionally, there is a tradeoff between selection and crossover. Greater selection pressures allow greater crossover rates. Analogically, smaller selection rates allow only smaller crossover rates. The tradeoff can be formalized using a simplification of the schema theorem, which claims that the expected number of copies of each partial solution or schema after selection and crossover is given by
$
s (1-\epsilon),
$
where $s$ characterizes the selection strength as a factor by which the number of best solutions will grow, and $\epsilon$ represents disruption of the schema by crossover. By ignoring mutation and making a conservative assumption that crossover always disrupts the schema, it is easy to show that $\epsilon=p_c$ and that setting $s=4$ and $p_c=0.5$ ensures net growth of 2 for the schemata contained in the best solution. 


The original parameter-less GA did not consider mutation. To incorporate mutation, the probability of disrupting a schema due to mutation would have to be incorporated into $\epsilon$. For a bit-flip mutation and binary strings, where each bit of a solution is flipped with a fixed probability $p_m$, a bounding case could assume that the schema under consideration spans across the entire solution. In that case, the probability of disrupting the schema due to mutation can be computed as $1-(1-p_m)^n$, where $n$ is the total number of bits in the solution.

For interacting variables where traditional variation operators fail, even the parameter-less GA is going to suffer from excessive disruption and ineffective mixing of building blocks, as the negative effects of blind variation cannot be eliminated by tweaking GA parameters, but only by modifying the operators themselves.

\subsection{Eliminating population size}
To eliminate the population size, the parameter-less GA simulates a collection of populations of different size~\cite{Harik:99a}. It is important that for any population size $N$, there exists a population of size greater or equal than $N$ in the collection. Otherwise, problems that require population size greater or equal than $N$ could not be solved. Consequently, the collection must contain infinitely many populations, the size of which cannot be upper bounded. 

The parameter-less GA arranges the collection of populations as a {\em sequence}. The size of the first population in the sequence is set to a small constant called the base population size. The size of the second population is twice the size of the first population. In general, each next population is twice the size of the previous population and the population size thus grows exponentially starting with the base population size. 

Each population in the collection is allowed to run one generation for each $k$ generations of the population twice as large, where $k\geq 2$ is an integer constant. The original parameter-less GA considered $k=4$. That means, that the smaller population was allowed to proceed at twice the speed of the next larger population, where speed is again measured with respect to fitness evaluations. It can be shown that for any $k\geq 2$, the infinite collection of populations can be simulated tractably without increasing the number of function evaluations until convergence by more than a logarithmic factor with respect to the optimal population size~\cite{Pelikan:99c}.

\subsection{How many generations for each population?}

Based on convergence theory for both large and small populations~\cite{Muhlenbein:93c,Thierens:94,Ceroni:01}, the number of generations can be assumed to be upper-bounded by a constant that does not depend on the population size. Since small populations process generations faster than larger populations, it seems reasonable to terminate simulation of each population at some point and use computational resources more efficiently.

The parameter-less GA terminates a population in the collection if either of the following criteria is satisfied:
\begin{itemize}
\item The population converges and it consists of many copies of a single solution. In this case, it can be expected that no more improvement will take place anymore or that the search will become very inefficient. Clearly, this criterion is not going to have much effect if niching is used.
\item A larger population has a greater average fitness. Since larger populations converge generally at most as fast as smaller populations, this situation indicates that the smaller population got stuck in a local optimum and it can thus be terminated. 
\end{itemize}

It is important to note that the logarithmic overhead computed in~\cite{Pelikan:99c} does not consider the termination criterion and the parameter-less GA should thus perform well even without terminating any populations in the collection.


\section{Parameter-less hBOA}
\label{section-parameter-less-hBOA}

The parameter-less hBOA incorporates the population-sizing technique of the parameter-less GA into hBOA. Since hBOA ensures growth and mixing of important building blocks via learning and sampling a probabilistic model of promising solutions, there is no reason to restrict crossover rate and all offspring can be created by sampling the probabilistic model of promising solutions. Additionally, any selection pressure that favors best candidate solutions can be used without changing scalability of hBOA by more than a constant factor. 

The parameter-less hBOA simulates a collection of populations 
\[P=\{P_0, P_1, \ldots \}.\] 
The size of the first population $P_0$ is denoted by $N_0$ and it is called the base population size. The size of the population $P_i$ is denoted by $N_i$ and it can be obtained by multiplying the base population size by a factor of $2^i$:
\[
N_i = N_{0} 2^i.
\]

For all $i\in\{1, 2, \ldots\}$, one generation of $P_i$ is executed after executing $k\geq 2$ generations of $P_{i-1}$. Here we use $k=2$ so that all populations proceed at the same speed with respect to the number of evaluations. Each population is initialized just before its first iteration is executed. The pseudocode that can be used to simulate the collection of populations described here is shown in Figure~\ref{figure-plhBOA}. This implementation is slightly different from the one based on a $k$-ary counter described in the first parameter-less GA study~\cite{Harik:99a}.

\begin{figure}
\begin{verbatim}
Parameter-less hBOA
  initialize P[0];
  generation[0]=0;
  max_initialized=0;
  i=0;
  while (not done) {
    simulate one generation of P[i];
    generation[i] = generation[i]+1;
    if (generation[i] mod k = 0) {
      i = i + 1;
      if (i>max_initialized) {
        initialize P[i];
        max_initialized=i;
      }
    }
    else
      i = 0;
  };
\end{verbatim}
\caption{The pseudocode of the parameter-less hBOA.}
\label{figure-plhBOA}
\end{figure}

\subsection{When to terminate a population?}

The same termination criteria as in the parameter-less GA can be used in the parameter-less hBOA. However, since hBOA uses niching, no population in the collection can be expected to converge for a long time. Additionally, we terminate each population after it executes for a number of generations equal to the number of bits in the input string. According to our experience, enabling each population to run for more generations does not improve performance further, whereas decreasing the limit on the number of generations might endanger convergence on exponentially scaled and hierarchical problems. Similarly as in the parameter-less GA, in the parameter-less hBOA populations can also be run indefinitely without increasing the worst-case overhead compared to the case with an optimal population size as predicted by theory.


\section{Experiments}
\label{section-experiments}

This section describes experimental methodology, test problems, and experimental results.

\subsection{Experimental methodology}

The parameterless hBOA was applied to artificial hierarchical and nearly decomposable problems and 2D $\pm J$ spin glasses with nearest neighbor interactions and periodic boundary conditions. For artificial problems, problem size was varied to examine scalability of the parameter-less hBOA. The performance of the parameter-less hBOA was compared to that of hBOA with optimal population size. For spin glasses, systems of different size were tested and 100 random instances were examined for each problem size to ensure that the results would provide insight into hBOA performance on a wide range of spin glass instances. To improve hBOA performance on spin glasses, a deterministic local search that flips each bit in a candidate solution until the solution cannot be improved anymore is used to improve each candidate solution before it is evaluated. The local searcher favors the best change at each iteration~\cite{Pelikan:thesis}. In all experiments, the base population size $N_{0}=10$ is used.

For each problem instance and each problem size, the parameter-less hBOA is first run to find the optimum in 100 independent runs and the total number of evaluations in every run is recorded. The average number of function evaluations is then displayed. These results are compared to the results for hBOA with the minimum population size that ensures that 30 independent runs converge to the optimum. The minimum population size was determined using bisection until the width of the resulting interval is at most $10\%$ of the lower bound.

The remainder of this section discusses test problems and experimental results.


\subsection{Deceptive function of order 3}
In the deceptive function of order 3~\cite{Ackley:87b,Deb:94b}, the input string is first partitioned into independent groups of $3$ bits each. This partitioning should be unknown to the algorithm, but it should not change during the run. A 3-bit deceptive function is applied to each group of 3 bits and the contributions of all deceptive functions are added together to form the fitness. The 3-bit deceptive function is defined as follows:
\begin{equation}
dec_3(u) = 
\left\{
\begin{array}{ll}
1 & \mbox{~~if $u=3$} \\
0 & \mbox{~~if $u=2$} \\
0.8 & \mbox{~~if $u=1$} \\
0.9 & \mbox{~~if $u=0$} 
\end{array}
\right.,
\end{equation}
where $u$ is the number of $1$s in the input string of $3$ bits. The 3-bit deceptive function is fully deceptive~\cite{Deb:91c}, which means that variation operators should not break interactions between bits in each group, because all statistics of lower order lead the algorithm away from the optimum. That is why most crossover operators as well as the model in UMDA will fail at solving this problem faster than in exponential number of evaluations, which is just as bad as with random search.  Since deceptive functions bound a broad class of decomposable problems, performance of the parameter-less hBOA on this class of problems should indicate what performance can be expected on other decomposable problems.

\begin{figure}[t]
\begin{center}
\epsfig{file=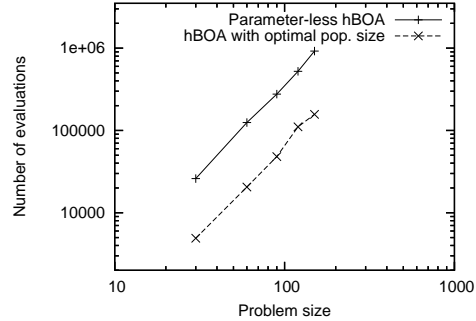,width=2.5in}
\end{center}
\caption{Parameter-less hBOA and hBOA with the optimal population size on the deceptive function of order 3.}
\label{figure-results-deceptive}
\end{figure}

Figure~\ref{figure-results-deceptive} shows the number of evaluations of the hierarchical BOA with optimal population size determined by the bisection method and the parameter-less hBOA on order-$3$ deceptive functions of $n=30$ to $n=150$ bits. The results indicate that the number of function evaluations for decomposable problems increases only by a near constant factor, and eliminating the parameters of hBOA thus does not affect scalability of hBOA on decomposable problems qualitatively.

\section{Hierarchical traps of order 3}

In an order-3 hierarchical trap with $L$ levels, the total number of bits in a candidate solution is assumed to be an integer power of 3, $n=3^L$. A candidate solution is evaluated on multiple levels and the overall value of the hierarchical trap is computed as the sum of contributions on all levels. On each level, a basis function similar to the deceptive function of order 3 is used:
\begin{equation}
basis(u,f_{lo},f_{hi}) = 
\left\{
\begin{array}{ll}
f_{hi} & \mbox{~~if $u=3$} \\
f_{lo}-u \frac{f_{lo}}{u-1} & \mbox{~~otherwise} 
\end{array}
\right.,
\end{equation}
On the lowest level, the candidate solution is partitioned into independent groups of 3 bits each, and each group is evaluated using the basis function with $f_{lo}=f_{hi}=1$. The contributions of $000$ and $111$ are thus equally good, and $000$ and $111$ are superior to other combinations of $3$ bits in any group. The contributions of all these groups are added together to form the overall contribution of the first level. Each group is then mapped (or interpreted) into a single symbol on the next level using the following interpretation function:
\begin{equation}
map(X) = 
\left\{
\begin{array}{ll}
0 & \mbox{~~if $X=000$}\\
1 & \mbox{~~if $X=111$}\\
- & \mbox{~~otherwise}
\end{array}
\right.,
\end{equation}
where $X$ denotes the input $3$ bits or symbols, and ``--'' is a null symbol. The second level thus contains $\frac{n}3$ symbols from $\{0, 1, -\}$. 

Symbols on the second level are also partitioned into independent groups of 3 bits each and each group contributes to the fitness on this level using the same basis function as on the first level, where $f_{hi}=f_{lo}=1$. However, groups that contain the symbol ``--'' do not contribute to the fitness at all. The overall contribution of the second level is multiplied by $3$ and added to the contribution of the first level. Each group is then mapped to the third level using the same interpretation function as was used to map the first level to the second one. 

The same principle is used to evaluate and map each higher level except for the top level, which contains $3$ symbols (the last 3 symbols are not mapped anymore). The contribution of $k$th level is always multiplied by $3^{k-1}$ so that the overall contribution of each level is of the same magnitude. The only difference when evaluating the top level of $3$ symbols is that $f_{lo}=0.9$, whereas $f_{hi}=1$.

The global optimum of hierarchical traps is in the string of all $1$s. However, blocks of $0$s and $1$s seem to be equally good on each level. Furthermore, any evolutionary algorithm is biased to solutions with many $0$s, because the neighborhood of solutions with many $1$s is inferior. Since hierarchical traps bound a broad class of hierarchical problems, performance of the parameter-less hBOA on this class problems should indicate what performance can be expected on other hierarchical problems.

\begin{figure}[t]
\begin{center}
\epsfig{file=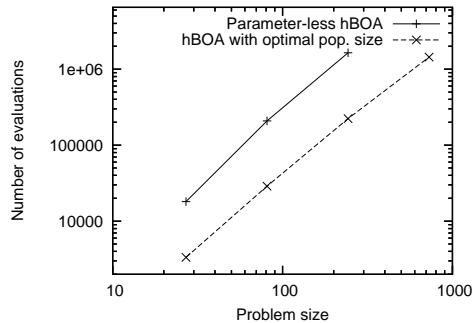,width=2.5in}
\end{center}
\caption{Parameter-less hBOA and hBOA with the optimal population size on the hierarchical trap.}
\label{figure-results-htrap}
\end{figure}

Figure~\ref{figure-results-htrap} shows the number of evaluations of the hierarchical BOA with optimal population size determined by the bisection method and the parameter-less hBOA on hierarchical traps of $n=27, 81, \mbox{and~} 243$ bits. The results indicate that the number of function evaluations for hierarchically decomposable problems increases only by a constant factor, and eliminating the parameters of hBOA does not affect scalability of hBOA on this class of problems qualitatively.

\section{2D Ising $\pm J$ spin glasses}

A 2D spin-glass system consists of a regular 2D grid containing $n$ nodes. The edges in the grid connect nearest neighbors. Additionally, edges between the first and the last element in each dimension are added to introduce periodic boundary conditions. 

With each edge, there is a real-valued constant associated with it called also a coupling constant. Each spin can obtain two values: $+1$ or $-1$. Given a set of coupling constants and spins, the energy of the spin glass system can be computed as
\begin{equation}
\label{equation-spin-glass-energy}
E(S) = \sum_{(i,j)\in E} s_i J_{i,j} s_j,
\end{equation}
where $i\in\{0, 1, \ldots, n-1\}$, $j\in\{0,1,\ldots,n-1\}$, $E$ is the set of edges in the system, $s_i$ and $s_j$ denote the values of the spins $i$ and $j$, respectively, and $J_{i,j}$ denotes the coupling constant between the spins $i$ and $j$. 

Here the task is to find the ground state for a 2D spin glass with specified coupling constants, where the ground state is the configuration of spins that minimizes the energy of the system given by Equation~\ref{equation-spin-glass-energy}. Each spin glass configuration is represented by a string of bits where each bit corresponds to one spin: a $0$ represents a spin $-1$, and a $1$ represents a spin $+1$. We created 100 random spin glasses of sizes $6\times 6$ ($36$ spins) to $14\times 14$ ($196$ spins) by generating coupling constants to be $+1$ or $-1$ with equal probabilities. Spin glasses with coupling constants restricted to $+1$ and $-1$ are called $\pm J$ spin glasses.

\begin{figure}[t]
\begin{center}
\epsfig{file=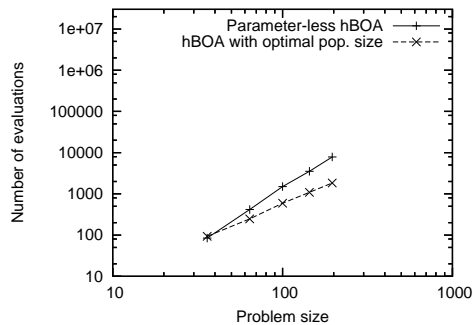,width=2.5in}
\end{center}
\caption{Parameter-less hBOA and hBOA with the optimal population size on $\pm J$ spin glasses.}
\label{figure-results-spin-glass}
\vspace*{-2ex}
\end{figure}

Figure~\ref{figure-results-htrap} shows the number of evaluations of the hierarchical BOA with optimal population size determined by the bisection method and the parameter-less hBOA on 2D $\pm J$ spin glasses. The good news is that the number of evaluations until convergence of the parameter-less hBOA still grows as a low-order polynomial with respect to the number of decision variables. The bad news is that, unlike for single-level and hierarchical traps, in this case the parameter-less population sizing does influence scalability of hBOA. More specifically, the order of the polynomial that approximates the number of evaluations increases by about~1. 

Clearly, a linear factor by which the number of evaluations increases disagrees with the existing theory, which claims that the factor should be at most logarithmic~\cite{Pelikan:99c}. Our hypothesis why this happens is that in this case the dynamics of hBOA with local search changes when increasing the population size. Using the hybrid method enables the two searchers---hBOA and the local searcher---to cooperate and solve the problem faster. hBOA allows the local searcher to exit the enormous number of local optima, which make local search by itself intractable. On the other hand, the local searcher decreases the population sizes in hBOA significantly by making the algorithm to focus on regions with local optima. The performance of the hybrid then depends on how well the division of labor between the local and global searcher is done. Increasing the population size in hBOA with local search can influence the division of labor in the hybrid. Since the parameter-less hBOA simulates a number of populations of different sizes, we believe that the division labor is affected and more opportunities are given to hBOA at the expense of the local searcher. We are currently verifying the above hypothesis.


\vspace*{-1ex}
\section{Summary and conclusions}
\label{section-conclusions}

This paper described, implemented, and tested the parameter-less hierarchical BOA. The parameter-less hBOA enables the practitioner to simply plug in the problem into hBOA without requiring the practitioner to first estimate an adequate population size or other problem-specific parameters. Despite the parameter-less scheme, low-order polynomial time complexity of hBOA on broad classes of optimization problems is retained. The parameter-less hBOA is thus a true black-box optimization algorithm, which can be applied to hierarchical and nearly decomposable problems without setting any parameters whatsoever. An interesting topic for future work is to develop techniques to improve hBOA performance via automatic tuning of other parameters, such as the maximum order of interactions in the probabilistic model or the window size in RTR.


\vspace*{-1ex}
\section*{Acknowledgments}

Martin Pelikan was supported by the Research Award at the University of Missouri at St. Louis and the Research Board at the University of Missouri at Columbia.

\bibliographystyle{splncs}
\bibliography{mybib}

\end{document}